%% file: eswa2026-fps-arxiv.tex
\documentclass{article}

\usepackage{arxiv}

\usepackage[utf8]{inputenc} 
\usepackage[T1]{fontenc}    
\usepackage{hyperref}       
\usepackage{url}            
\usepackage{booktabs}       
\usepackage{amsfonts}       
\usepackage{nicefrac}       
\usepackage{microtype}      
\usepackage{lipsum}         
\usepackage{graphicx}
\usepackage{doi}

\usepackage{graphicx}%
\usepackage{multirow}%
\usepackage{amsmath,amssymb,amsfonts}%
\usepackage{amsthm}%
\usepackage{mathrsfs}%
\usepackage[title]{appendix}%
\usepackage{xcolor}%
\usepackage{textcomp}%
\usepackage{manyfoot}%
\usepackage{booktabs}%
\usepackage{algorithm}%
\usepackage{algorithmicx}%
\usepackage{algpseudocode}%
\usepackage{listings}%

\usepackage{cleveref}
\usepackage{subcaption}
\usepackage[export]{adjustbox}
\usepackage[english]{babel}

\newcommand{\ProjectArena}{\mbox{\textit{Project Arena}}}
\newcommand{\CubeTwo}{\mbox{\textit{Cube 2}}}
\newcommand{\CubeTwoS}{\mbox{\textit{Cube 2: Sauerbraten}}}
\newcommand{\SMT}{Spatial-Layout}

\hypersetup{
	pdftitle={Procedural Generation of First Person Shooter Maps using Map-Elites},
	pdfsubject={q-bio.NC, q-bio.QM},
	pdfauthor={Simone de~Donato, Pier~Luca Lanzi, Daniele Loiacono},
	pdfkeywords={First keyword, Second keyword, More},
}

\title{Procedural Generation of First Person Shooter Maps using Map-Elites}

\newif\ifuniqueAffiliation
\uniqueAffiliationtrue

\ifuniqueAffiliation 
\author{ \href{https://orcid.org/0009-0002-1844-9845}{\includegraphics[scale=0.06]{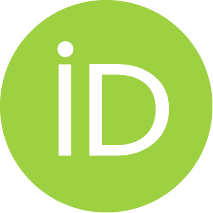}\hspace{1mm}Simone de~Donato}\\
	Politecnico di Milano --- DEIB\\
	\texttt{simone.dedonato@mail.polimi.it} \\
	\And
	\href{https://orcid.org/0000-0002-1933-7717}{\includegraphics[scale=0.06]{orcid.pdf}\hspace{1mm}Pier Luca Lanzi} \\
	Politecnico di Milano --- DEIB\\
	\texttt{pierluca.lanzi@polimi.it} \\
	\And
	\href{https://orcid.org/0000-0002-5355-0634}{\includegraphics[scale=0.06]{orcid.pdf}\hspace{1mm}Daniele Loiacono} \\
	Politecnico di Milano --- DEIB\\
	\texttt{pierluca.lanzi@polimi.it} \\
}
\else
\usepackage{authblk}

\newbox{\orcid}\sbox{\orcid}{\includegraphics[scale=0.06]{orcid.pdf}} 
\author[1]{%
	\href{https://orcid.org/0000-0000-0000-0000}{\usebox{\orcid}\hspace{1mm}David S.~Hippocampus\thanks{\texttt{hippo@cs.cranberry-lemon.edu}}}%
}
\author[1,2]{%
	\href{https://orcid.org/0000-0000-0000-0000}{\usebox{\orcid}\hspace{1mm}Elias D.~Striatum\thanks{\texttt{stariate@ee.mount-sheikh.edu}}}%
}
\affil[1]{Department of Computer Science, Cranberry-Lemon University, Pittsburgh, PA 15213}
\affil[2]{Department of Electrical Engineering, Mount-Sheikh University, Santa Narimana, Levand}
\fi

\renewcommand{\headeright}{}
\renewcommand{\undertitle}{}

\begin{document}
	\maketitle
	
	\begin{abstract}
		\input{sections/abstract}
	\end{abstract}


%
%

\section{Introduction}
\input{sections/sec_introduction}

\section{Related Work}
\input{sections/sec_related_short}

\section{Project Arena}
\label{sec:project_arena}
\input{sections/sec_project_arena}

\section{FPS Map Representations}
\label{sec:representations}
\input{sections/sec_map_representations}
\section{FPS Map Features}
\label{sec:features_executive}
\input{sections/sec_features_executive}

\section{Experimental Results}
\input{sections/sec_experiments_gecco}

\section{Conclusions}
\input{sections/sec_conclusions}

\bibliographystyle{elsarticle-num}
\bibliography{bibliography/ei2026-fps,bibliography/ei2026-fps-clear}
\end{document}

%% file: sections/abstract.tex
We investigate the application of MAP-Elites (a well-known quality diversity algorithm) to design levels for First-Person Shooter (FPS) games. We consider two well-known map representations (All-Black and Grid-Graph) and introduce two novel representations (Point-Line and \SMT) that improve the characterization of FPS maps. We define a series of metrics to describe maps' topological properties (which solely depend on maps' layout),  and emergent properties (which must be evaluated through actual gameplay). We perform an in-depth analysis to identify the most suitable features to guide MAP-Elites illumination process. We apply MAP-Elites with Sliding Boundaries (MESB) to evolve populations of FPS maps. Our results show that the new representations can generate maps with higher diversity and quality than the representations previously used for evolving FPS maps.

%% file: sections/sec_introduction.tex
Search-based procedural content generation has been applied to evolve maps for FPS games using traditional \cite{cardamone_evolving_2011, lanzi_evolving_2014, DBLP:conf/aiide/CachiaLY15,loiacono_fight_2017} and interactive evolutionary computation \cite{olsted_interactive_2015}. All these approaches focus on one specific aspect of level design, such as the average gameplay time between kills \cite{cardamone_evolving_2011}, gameplay balance \cite{lanzi_evolving_2014,DBLP:conf/aiide/CachiaLY15}, enabling fleeing behavior \cite{loiacono_fight_2017}, or adherence to a set of design guidelines \cite{olsted_interactive_2015}. These approaches generate maps that maximize one specific objective and usually explore a limited design space \cite{olsted_interactive_2015}.

In this paper, we explore applying quality diversity methods, specifically MAP-Elites \cite{mouret_illuminating_2015}, to evolve maps for multiplayer FPS games. 
MAP-Elites is an evolutionary algorithm for exploring and optimizing diverse, high-performing solutions across an illuminated search space defined by a pair of features. Individuals are characterized by a fitness that the system must maximize and an archive of elite solutions that the algorithm maintains. The solutions are stored in bins identified by the values of the features the system uses to illuminate the search space. In particular, we apply MAP-Elites with Sliding Boundaries (MESB) \cite{fontaine_mapping_2019}, which maintains an archive of adaptive bins.
We consider two well-known map representations (All-Black and Grid-Graph) \cite{cardamone_evolving_2011} and introduce two new representations (Point-Line and \SMT) we specifically created to enrich the characterization of FPS maps. We define several metrics to describe maps' topological properties, computed from the maps' layout, and emergent properties, evaluated through actual gameplay. 
We performed an extensive analysis of the available features and focused on the ones that, in the literature, are regarded as the most interesting ones for FPS map design \cite{DBLP:conf/fdg/HullettW10,cardamone_evolving_2011,DBLP:conf/cig/BallabioL19}. 
In particular, we focus on three topological features: the walkable \textit{area} and the map symmetry (\textit{maxSymmetry}), which provide high-level descriptions of the map topology \cite{DBLP:conf/fdg/HullettW10,cardamone_evolving_2011}; and the map eccentricity (\textit{averageEccentricity}), which was specifically introduced for graph-based representations of FPS maps \cite{ DBLP:conf/cig/BallabioL19}. We also consider the map \textit{pace}, which measures the frequency of combat engagements \cite{cardamone_evolving_2011}.\footnote{We refer the reader to \cite{denodato:2024:thesis} a comprehensive discussion of all the features we analyzed.}
We combined the four features into two pairs: \textit{area-maxSymmetry}, capturing only topological characteristics of FPS maps, and \textit{pace-averageEccentricity}, combining gameplay and topological features. We compared the maps evolved using the four representations with MAP-Elites using the two feature pairs. The results show that the new representations can generate maps with higher diversity and overall quality than representations previously used for evolving FPS maps. All the experiments were run using an extension of the open source framework \ProjectArena\ \cite{ProjectArena} for research in first-person shooter games and the PyRibs library \cite{tjanaka_pyribs_2023} for quality diversity optimization. All the code to run the experiments discussed in this paper and the experimental data presented are available at \cite{duplication_github}.

%% file: sections/sec_related_short.tex
Search-based Procedural Content Generation (SB-PCG) \cite{togelius_search-based_2010} applies a generate-and-test approach to procedural content generation. Candidates are first generated, then evaluated using a fitness function, and new candidates are created based on the result of the evaluation. SB-PCG encompasses several methods of genetic and evolutionary computation and it has been applied to several game genres including first-person shooters.

\subsection{SB-PCG for FPS Games}
\label{ssec:sp-pcg-fps}
\input{sections/ssec_sbpcg_fps.tex}
\subsection{Map-Elites for Games}
\label{ssec:qd_pcg}
\input{sections/ssec_qd_fps}

%% file: sections/ssec_sbpcg_fps.tex
Cardamone et al. \cite{cardamone_evolving_2011} were the first to apply search-based techniques to generate playable FPS maps for the ``deathmatch" game mode of \CubeTwoS\ \cite{cube2}. In their work, they introduced four genotypes (\textit{All-White}, \textit{All-Black}, \textit{Grid} and \textit{Random-Digger})
and the mapping functions to generate the corresponding maps for \CubeTwo. 
Candidate maps were evaluated by simulating a 10 minutes match with four bots; fitness was computed as the average time between the bot starts fighting until the time it is killed by an opponent. 
In \cite{cardamone_evolving_2011}, the authors argued that interesting maps 
should have features allowing for long fights to happen, such as escape routes and well-placed resources. 
Lanzi et al. \cite{lanzi_evolving_2014} focused on evolving maps that allowed for balanced 1-vs-1 "deathmatch" matches between bots with different skill levels and play style (e.g., aggressive, defensive). They also used \CubeTwo, the \textit{All-Black} genotype \cite{cardamone_evolving_2011}, and evaluated maps through simulation of matches involving different types of bots. Fitness function was computed as the \textit{entropy} of the score (number of kills) ratio distribution
since the higher the entropy, the more balanced the score distribution, and therefore the more balanced the match.
Bhojan and Wong \cite{bhojan_arena_2014} focused on faster evolution methods to evolve maps \textit{online} for the ``Capture the Flag" game mode 
in a game they developed to test their method. Bhojan and Wong \cite{bhojan_arena_2014} aimed at fast evolution and therefore did not use simulations but
defined a fitness which was the sum of measured map features such as, connectivity, 
forced collision points, flag fairness, overall flag fairness \cite{bhojan_arena_2014}.
They generated maps by placing tiles that were either an indoor area, an outdoor area, or inaccessible, following a rule of adjacency with pre-existing tiles. 
The maps were then cleaned of artifacts and scanned with a \textit{Flood Fill} algorithm \cite{enwiki:1257255207} to identify regions that would then be connected with doors where needed. 
The maps were finally decorated with strategic points such as spawn points, flag locations and covers. 
%
%
Ølsted et al. \cite{olsted_interactive_2015} moved to evolving maps for the game \textit{FPSEvolver} focusing on the ``Bomb Defusal" game mode, similar to that of \textit{Counter-Strike}. 
%
They defined a set of guidelines (\textit{the good engagement}) and designed a genotype that could create layouts that adhere to these concepts, avoiding dead ends and ensuring arenas. 
They applied \textit{iterative evolutionary computation} to evolve maps with human subjects that played the maps
and then were asked to leave simple binary (thumb-up or thumb-down) feedback which was used to guide the evolution \cite{olsted_interactive_2015}.
Cachia et al. \cite{DBLP:conf/aiide/CachiaLY15} evolved maps that occupied more than one floor in \CubeTwo\ with a genome
that combined different representations for the first and the second floor---the former as a set of corridors and connected rooms
identifying passable tiles, the latter using a \textit{Random-Digger} \cite{cardamone_evolving_2011}.
Fitness was based on simulations with artificial agents, on the available space for navigation, and on the presence of specific level patterns.
Loiacono et al. \cite{loiacono_fight_2017} took an approach similar to \cite{cardamone_evolving_2011}, 
leveraging the same \textit{All-Black} representation to evolve maps that foster fleeing behavior. 
Fitness was computed as the number of times during a match that a bot lost track of an enemy it is currently fighting. 
They showed that their approach could generate maps that enabled fleeing behavior in bots that are designed 
to always attack while creating maps that significantly reduce pace \cite{loiacono_fight_2017}.

%% file: sections/ssec_qd_fps.tex
\textit{MAP-Elites} has been used to generate content for a variety of game genres such as, platformers, 
	``Zelda-like" adventures, puzzles, and card games.
%

\medskip\noindent\textbf{Platformer Games.}
Khalifa et al. \cite{khalifa_intentional_2019} introduced \textit{Constrained MAP-Elites} to evolve a variety of playable \textit{Super Mario Bros.} levels and compared different simulation approaches to generate them \cite{khalifa_intentional_2019}. Warriar et al. \cite{warriar_playmapper_2019} implemented \textit{PlayMapper}, a variation on the MAP-Elites algorithm, that is able to generate \textit{Super Mario Bros.} levels of different sizes and grants a significant amount of control over their generation. 
Fontaine et al. \cite{fontaine_illuminating_2021} applied a set of \textit{MAP-Elites} methods to illuminate the latent space of a \textit{Generative Adversarial Network (GAN)} assessing the algorithms' performance on the task of generating scenes for \textit{Super Mario Bros.} with specific features.

\medskip\noindent\textbf{Adventure Games.}
Alvarez et al. \cite{alvarez_empowering_2019} introduced \textit{Interactive Constrained MAP-Elites} 
	to evolve dungeon rooms in \textit{Evolutionary Dungeon Designer} (EDD), 
	a mixed-initiative co-creativity tool. 
They combined EDD with \textit{Feasible-Infeasible 2-Population} (FI2Pop) genetic algorithm \cite{kimbrough_feasibleinfeasible_2008} 
	to create an interactive version of \textit{Constrained MAP-Elites} in which users guided the search by selecting the features to explore and gave feedback on the generated rooms \cite{alvarez_empowering_2019}.
Charity et al. \cite{charity_mech-elites_2020} searched for levels with specific mechanics in the \textit{GVG-AI Framework}'s games \cite{perez-liebana_general_2019}, 
	which include a ``Zelda-like" game \cite{charity_mech-elites_2020}. 
González-Duque et al. \cite{gonzalez-duque_finding_2020} applied \textit{Intelligent Trial-and-Error} (a form of Bayesian Optimization that uses MAP-Elites to generate the priors) to perform \textit{Dynamic Difficulty Adjustment}, producing levels of the desired difficulty for a ``Zelda-like" game.
Viana et al. \cite{viana_illuminating_2022} applied MAP-Elites to generate levels with locked doors missions and enemy placement, 
	with high quality and diversity .

\medskip\noindent\textbf{Puzzle Games.}
Charity et al. \cite{charity_baba_2020} developed \textit{Baba is Y'all}, a mixed-initiative version of \textit{Baba is You}
where players design levels in collaboration with the machine. The game allows players to edit \textit{Baba is You} levels, similar to \textit{Super Mario Maker},
with the added goal of filling the archive of the underlying MAP-Elites algorithm, where levels are placed based on the combination of mechanics (rules) that are activated when solving them. The system can propose various starting levels for the user to edit based on what is already in the archive and what's currently missing to help players fill the archive \cite{charity_baba_2020}. 

\medskip\noindent\textbf{Card Games.}
MAP-Elites has also been used to generate competitive card decks for \textit{Hearthstone}. Fontaine et al. \cite{fontaine_mapping_2019} noted that \textit{MAP-Elites} uniformly divides the behavior space, however this can lead to a mismatch with the true distribution of the behavior space, which would result in a less effective illumination. Moreover, they argue that knowing \textit{a priori} distribution can be hard or even impossible. To solve these issues, they propose \textit{MAP-Elites with Sliding Boundaries} (MESB), a variation of MAP-Elites where boundaries between cells are not placed uniformly based on the value of the features, but instead are dynamically placed at certain percentage marks of the distribution. A remap frequency is specified, meaning that the boundaries are periodically recalculated, allowing for a good estimation of the distribution of the search space. \cite{fontaine_mapping_2019}

\medskip\noindent\textbf{Other Genres.}
Khalifa et al. \cite{khalifa_talakat_2018} have applied \textit{MAP-Elites} to generate levels for bullet hell games. 
Gravina et al. \cite{DBLP:conf/cig/GravinaLY16} applied Constrained Surprise Search to generate weapons for Unreal Tournament III first person shooter.

\medskip
Quality Diversity methods have yet to be applied to the generation of levels for FPS games, which is the goal of our research.

%% file: sections/sec_project_arena.tex
\ProjectArena\ \cite{ProjectArena} is an open-source platform for research in first-person shooter games 
	developed using the Unity game engine. It has a modular architecture structured 
	in a series of managers, bots, and visualizations that can be easily customized, 
	thanks to the many available Unity extensions.

\medskip\noindent\textbf{Managers.} The \textit{Game Manager} handles the main lifecycle, including setup, game modes, matches, map management, and generation. \ProjectArena\ has three game modes: \textit{Duel}, implementing the classic deathmatch available in all FPSs; \textit{Target Rush}, with increasingly difficult waves of attacking enemies that the player must eliminate; and \textit{Target Hunt}, in which the player must search and eliminate enemies within a time limit. The \textit{Map Manager} handles the load/save of existing maps as well as the generation of maps. It currently provides three classical map generation methods based on cellular automata, random digger, and binary space partition \cite{shaker:2016:pcg}. The \textit{Experiment Manager} lets users organize a series of experiments in a collection of studies, each one consisting of a set of cases (defined by a set of game modes, maps, and bots) and an evaluation procedure (round-robin by default); all the information generated during the experiments are logged; designers can also include an optional survey if the experiments involve human subjects.

\medskip\noindent\textbf{Bots.} The framework provides several bots that 
	can be customized based on their (i) skill level, expressed as a percentage between 0\% (rookie) and 100\% (expert); 
	(ii) fighting strategy, which depends on their weapon (e.g., a sniper rifle needs a completely different tactics than a shotgun or grenade launcher).

%% file: sections/sec_map_representations.tex
In our study we considered four representations of maps for first-person-shooters:
\textit{All-Black} \cite{cardamone_evolving_2011}, 
	\textit{Grid-Graph} \cite{ProjectArena};
\textit{Point-Line} and \textit{\SMT} that we specifically designed for using with MAP-Elites. 

\medskip\noindent\textbf{All-Black} \cite{cardamone_evolving_2011} carves walkable spaces from a map that is initially filled with walls.
Maps are lists of triplets representing \textit{rooms} and \textit{corridors}. Rooms are defined by the coordinates of their left-bottom corner, height, and width; corridors are defined by the coordinates of their left-bottom corner and length, which encodes whether they are horizontal (positive length) or vertical (negative). This representation has issues with locality (small mutations in the genome can cause major changes in the map phenotype) and redundancy (different genomes can produce identical maps). Furthermore,  the resulting maps often contain several dead ends and confusing features \cite{olsted_interactive_2015}.

\medskip\noindent\textbf{Grid-Graph} is inspired by the dungeon generation in Rogue~\cite{enwiki:rogue}, the father of the rogue-like game genre. It starts with a map full of walls (like \textit{All-Black}) that it partitions into an $R\times C$ grid (fixed to $3\times 3$ in Rogue~\cite{enwiki:rogue}); each grid position can contain at most one room, defined by its position within the cell, its width, and height. Rooms in adjacent cells can be connected by corridors, which are encoded as a boolean value for each couple of rooms horizontally and vertically adjacent. This approach is more compact and less redundant than \textit{All-Black} however the number of layouts depends on the grid size and it is generally more limited than \textit{All-Black}. 

%

\medskip
\textit{All-Black} has known limitations \cite{olsted_interactive_2015} and \textit{Grid-Graph} allows for a limited number of topologies which depends on the predefined grid size. Accordingly, we introduced two representations we specifically designed for quality diversity search algorithms 
	by solving some of the limitations of \textit{All-Black} while allowing for a richer and much diverse search space. 

\medskip\noindent\textbf{Point-Line} genome is a sequence of $P$ tuples $\langle p_1, p_2, s_1, s_2, c\rangle$ each representing a pair of rooms connected by an L-shaped corridor; $p_1$ and $p_2$ are the rooms' positions; $s_1$ and $s_2$ are their sizes; $c$ represents the orientation of the L-shaped corridor.
%
%
Point-Line improves locality over \textit{All-Black} by explicitly connecting rooms with corridors. It eliminates the issue of dead ends of \textit{All-Black} \cite{olsted_interactive_2015} and it is more compact since one point represents both rooms' and corridors' coordinates. 
\textit{All-Black} rarely generates long corridors since these are built as a sequence of short corridors. In contrast, by defining corridors using their start and end positions, Point-Line gives long and short corridors an equal chance making it more likely to explore a wider variety of designs than \textit{All-Black}. However, it does not solve the issue of redundancy. 

\medskip\noindent\textbf{\SMT} is inspired to the work of \cite{whitehead_spatial_2020}. The genome consists of 
	(i) a list of $R$ axis-aligned rectangular rooms defined by their width and height;
	(ii) a list of $L$ line segments defined by two points;
	(iii) the minimum separation parameter $s$ which determines the minimum distance between rooms' borders in the genome. 
\SMT\ does not include room positions since these are determined using a Satisfiability Modulo Theories (SMT) solver, implemented using Python's Z3 library,
	based on the following constraints: rooms are inside the maps boundaries; rooms do not overlap; rooms are close enough to the lines.
Once room positions are determined, corridors are positioned by computing the minimum spanning tree from the Delaunay triangulation of the rooms, which ensures connectivity. Mutation changes rooms and lines parameters; crossover swaps rooms and lines between two parent genomes. 

Whitehead \cite{whitehead_spatial_2020} tackled dungeon designs 
	where an SMT solver would suitably represent a linear layout with a start, an end, and some optional areas. 
FPS maps have more complex topologies that entail loops, arenas, and alternate routes \cite{hullett_design_2010}.
To improve map complexity and navigability, we extended the approach of \cite{whitehead_spatial_2020} 
	with a heuristic to add corridors between rooms intersected by the genome's lines, if they were not already connected. 
%
Note that, SMT solver is not deterministic and the same genome can produce more topologies, which might lead to poor locality but more diversity. 
Furthermore, \SMT\ genomes may not allow a feasible solution and thus might be discarded. 

%% file: sections/sec_features_executive.tex
%
We defined several features to describe maps based on their topology (46 features) and on the gameplay they enable (23 features). In this section, we focus on the most relevant features 
that we also discuss in the experimental section; we refer the interested reader to \cite{denodato:2024:thesis} for the complete list of 69 features.


\medskip\noindent\textbf{Topological Features} describe the map's structure. They solely depend on the map's layout and are extracted from the phenotype without the need for simulations.
%
First, we apply a technique for region decomposition based on the line segment Voronoi diagrams to extract a graph representation of the map's topology. Then, we compute several metrics such as the number of loops, the average distance between rooms,  the number of alternative paths, etc. \cite{denodato:2024:thesis}.
We also apply  a grid-based method to extract more advanced features such as the \textit{area}, computed as the number of tiles that are walkable compared to the total number of tiles in the map, or the map's visibility matrix, describing for each tile the number of tiles that are visible from it. We gain insights on the average visibility, the number of local maxima, their average distance and the maximum visibility. 
Finally, we compute high-level map properties like the map symmetry as the percentage of tiles that are symmetrical counterpart with respect to an axis (\textit{xSymmetry} and \textit{ySymmetry}) and the maximum symmetry that the map has (\textit{maxSymmetry} computed as the max of \textit{xSymmetry} and \textit{ySymmetry}). Another example is the \textit{averageEccentricity} of the rooms' eccentricity that is computed, for each room, as the maximum distance between the room and all other rooms. 


\medskip\noindent\textbf{Emergent Features} are computed from actual gameplay sessions which, in our study, are simulated using bots implementing a variety of fighting strategies. 
They include metrics such as pace, average fight time, target loss rate, kill difference, etc. \cite{denodato:2024:thesis}. 
%
Pace measures the frequency of combat engagements, normalized between 0 and 1. It is a function of the number of fights occurred and the time between them, computed as,
\begin{equation}
pace = 2 \left(1 + \exp \left(-5 \dfrac{N_F}{\sum T_E}\right) \right)^{-1} - 1
\end{equation}
where $N_F$ is the number of fights and $T_E$ is the time to engage; the sigmoid function computes values close to 0.9 when the average time to engage is close to 3 seconds.


%% file: sections/sec_experiments_gecco.tex
We performed a series of experiments to evaluate the quality and diversity of the maps evolved using MAP-Elites and the four representations of FPS maps (Section~\ref{sec:representations}). At first, we performed a series of simulations with random maps to collect data about the several features we defined (Section~\ref{sec:features_executive}) and focused on the ones that, according to the literature, are the most relevant for FPS map design \cite{DBLP:conf/fdg/HullettW10,cardamone_evolving_2011,DBLP:conf/cig/BallabioL19}. 
In this section, we present the result of our analysis on the four most important features (\textit{area}, \textit{MaxSymmetry}, \textit{pace}, and \textit{averageEccentricity}) and refer the reader to \cite{denodato:2024:thesis} for an analysis of the other features we considered. 
We grouped the four features into two pairs: \textit{area} and \textit{MaxSymmetry}, which capture only topological characteristics of the maps; \textit{pace} and \textit{averageEccentricity}, which combine gameplay and topological characteristics. Next, we applied \textit{MAP-Elites with Sliding Boundaries} (MESB) \cite{fontaine_mapping_2019} to evolve highly diverse, high quality, FPS maps which could enable a balanced gameplay.

\subsection{Fitness}
\label{ssec:fitness}
Previous research in the evolution of FPS maps used fitness function either based on topological \cite{bhojan_arena_2014,olsted_interactive_2015} or emergent gameplay properties \cite{cardamone_evolving_2011,lanzi_evolving_2014,DBLP:conf/aiide/CachiaLY15,loiacono_fight_2017}. 
In our experiments, we defined a fitness inspired to the latter approaches \cite{cardamone_evolving_2011,lanzi_evolving_2014,DBLP:conf/aiide/CachiaLY15,loiacono_fight_2017} 
	that enables a balanced gameplay in a  \textit{Duel} (i.e., 1-vs-1 \textit{deathmatch}) game mode independently of the skill levels and the equipment of the players involved. 
We selected two bots with very different skill levels (15\% and 85\%) equipped with two weapons that require completely different fighting tactics: 
	(i) a sniper rifle, suited for long distance combat and more effective in environments with large visibility areas, 
	(ii) a shotgun, effective only in close combat scenarios. 
Our fitness is computed as the \textit{entropy of the match balance} averaged over five matches; 
	the entropy of a single match is defined as, 
\begin{equation}
	entropy = -\sum_{i=1}^{n}  \left(\dfrac{k_i}{k_{tot}}\right) \log_2 \left(\dfrac{k_i}{k_{tot}}\right)
\end{equation}
where $k_i$ is the number of kills, $k_{tot}$ is the total number of kills during the simulated match.
Entropy maximizes the game balance, the higher the entropy, the more balanced the match \cite{lanzi_evolving_2014,DBLP:conf/aiide/CachiaLY15}. 


\subsection{Design of Experiments}
\label{ssec:design}
We applied MAP-Elites with Sliding Boundaries (MESB) \cite{fontaine_mapping_2019} using an archive of 10 bins for each feature, for a total of 100 maximum solutions. Each run starts with 20 random individuals and lasts for 400 iterations using 10 \textit{emitters}; thus, each iteration generates 10 new individuals. Fitness is computed as the average entropy over five matches (Section~\ref{ssec:fitness}). After a map is evaluated using five matches, the illuminating features are computed for each match, averaged, and the resulting values are used to update the corresponding \textit{MESB} archive with the new individual. The experiments were performed using the PyRibs library \cite{tjanaka_pyribs_2023} for quality diversity optimization which is built on a highly modular conceptual QD framework and provides access to  all the most important algorithms and statistics. 

\subsection{\textit{area} and \textit{maxSymmetry}}
\label{ssec:area-maxsymmetry}
\input{sections/ssec_area_maxsymmetry}

\subsection{\textit{pace} and \textit{averageEccentricity}}
\label{ssec:pace-average-eccentricity}
\input{sections/ssec_pace_average_eccentricity}

%% file: sections/ssec_area_maxsymmetry.tex
In the first set of experiments, we applied  MAP-Elites with Sliding Boundaries (MESB) using \textit{area} and \textit{maxSymmetry} features to illuminate the search. Figure~\ref{fig:heatmap_exp1} reports the heatmaps of the archives generated for each representation. MESB dynamically adapts the archive's bins' boundaries based on the feature distribution \cite{fontaine_mapping_2019}; accordingly the four archives are defined over different areas of the search space but in Figure~\ref{fig:heatmap_exp1} we use the same axis ranges for comparison.
Figure~\ref{fig:area-maxsimmetry} reports  
(i) the maximum value of entropy achieved in the archive over time (\ref{fig:max_score_exp1});
(ii) the Complementary Cumulative Distribution Function (CCDF) of the fitness of elites (\ref{fig:ccdf_exp1}), 
	which evaluates the probability that a random elite has an \textit{entropy} above a certain level; 
(iii) the QD score of the archive overtime (\ref{fig:qd_score_exp1}),
	computed as sum of entropy of all elites in the archive;
(iv) the size of the archive over time (\ref{fig:size_exp1}).


%
%

\input{sections/fig_performance_exp1}
\input{sections/fig_visibility_exp1}

Figure~\ref{fig:heatmap_exp1}
 shows that the four representations illuminate the search space in different ways. \textit{Grid-Graph} generates by far the maps with the most amount of symmetry, stemming from its grid-like approach; however, the maps are too small and therefore uninteresting for human players \cite{hullett_design_2010}. \textit{Point-Line} produces the largest range of illuminating features especially for the \textit{area} with maps that are up to 50\% walkable. 

Figure~\ref{fig:max_score_exp1} shows that \textit{Grid-Graph} and \textit{Point-Line} can reach almost maximum \textit{entropy}, while \textit{All-Black} and \textit{\SMT} show steady increase but lower entropy values. 
The difference in \textit{entropy} is probably due to the maps these representations generate.  \textit{All-Black} and \textit{\SMT} struggle to create long corridors or large open spaces where a \textit{sniper} bot thrives; in contrast, \textit{Grid-Graph} and \textit{Point-Line} can easily generate such maps. 
The CCDF curves in Figure~\ref{fig:ccdf_exp1} show that \textit{Grid-Graph} generates several maps of very high entropy, while \textit{Point-Line} has a higher average entropy. \textit{Grid-Graph} has the lowest QD score while our representations Point-Line and \SMT\ outperform both representations reaching similar QD scores at the end. \textit{\SMT} has a slight edge over the others thanks to its fuller archive despite the lower average score of the elites (\ref{fig:qd_score_exp1}). Interestingly, \textit{\SMT} reaches the highest QD score earlier on but then the curve flattens, suggesting that this representation can quickly generate a wide variety of high performing maps, but it may struggle to improve them further. It is worth remembering that, unlike standard grid-like MAP-Elite archives \cite{mouret_illuminating_2015}, MESB archives consist of bins whose boundaries are periodically updated, which can cause sudden decrease in the QD score when such updates merge bins, lowering the number of elites in the archive.
Figure~\ref{fig:size_exp1} shows that the \textit{Grid-Graph} representation has the fewest elites in the archive, confirming the limited variety that this representation can generate. \textit{All-Black} achieves slightly better results, followed by Point-Line and then \SMT, which achieves the most elites in the archive. 
These results suggest that \SMT\ can generate more variety although \textit{Point-Line} achieves similar results with \textit{higher} area and \textit{maxSymmetry} values.

\medskip\noindent\textbf{Analysis of Maps in the Archive.} The map features we defined are proxies to elements that game designers consider important for FPS maps \cite{hullett_design_2010}. It is however infeasible to judge maps quality solely based on their features. Accordingly, we performed a visual analysis of the evolved maps to better characterize the results produced by the four representations. 
%
Figures \ref{fig:best_maps_ab_exp1}, \ref{fig:best_maps_point_exp1}, and \ref{fig:best_maps_smt_exp1}
	shows four examples of high fitness (entropy) maps evolved for the \textit{All-Black} (\ref{fig:best_maps_ab_exp1}), \textit{Point-Line} (\ref{fig:best_maps_point_exp1}), and \textit{\SMT} (\ref{fig:best_maps_smt_exp1}) representations. We did not include \textit{Grid-Graph} representation since it generated rather basic maps mainly consisting of large unique areas, see \cite{denodato:2024:thesis} for further experiments and maps examples.

All representations generate maps with large open spaces or long narrow corridors. Sniper bots tend to achieve an advantage when they are positioned at the edge of a map and are able to gain good visibility over most of the map. \textit{Point-Line} has fewer dead ends and chokepoints and when present they are implemented as corridors (e.g., \ref{fig:best_maps_exp1_pl2}, \ref{fig:best_maps_exp1_pl3}) which favor shotgun bots more than sniper bots.  
The layout of \textit{All-Black} maps (Figure~\ref{fig:best_maps_ab_exp1}) are noisy with many dead-ends, short corridors, and sharp turns, which favor shotgun bots; accordingly they tend to be unbalanced and thus with a lower entropy as already noted from Figure~\ref{fig:area-maxsimmetry}.
\SMT\ maps are a trade-off between \textit{All-Black} and \textit{Point-Line}: they are not as noisy and unbalanced as \textit{All-Black} while still having 
	varied layouts with some alternate paths and covers; they also have a higher size, and QD score than \textit{Point-Line} (Figure~\ref{fig:area-maxsimmetry}).

%% file: sections/fig_performance_exp1.tex
\begin{figure}[t]
	\centering
	\subfloat[All-Black]{\includegraphics[width=0.48\textwidth]{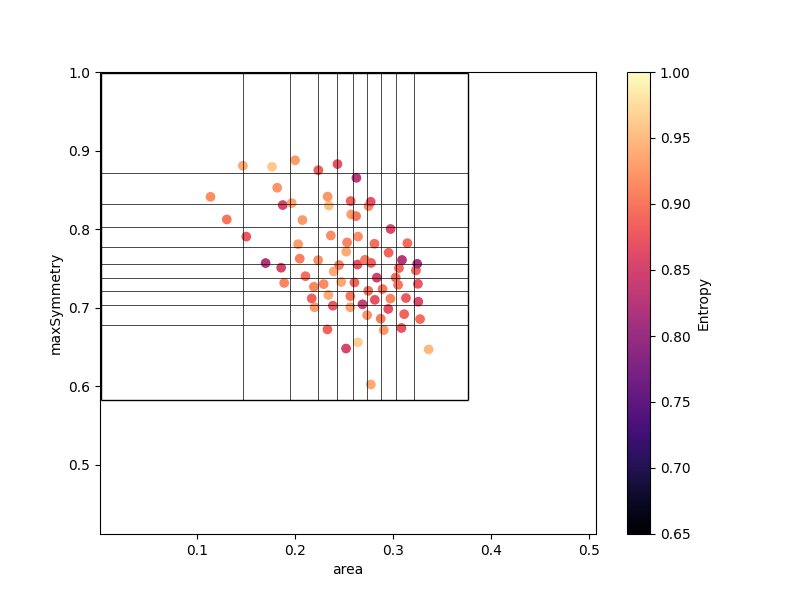}}
	\quad
	\subfloat[Grid-Graph]{\includegraphics[width=0.48\textwidth]{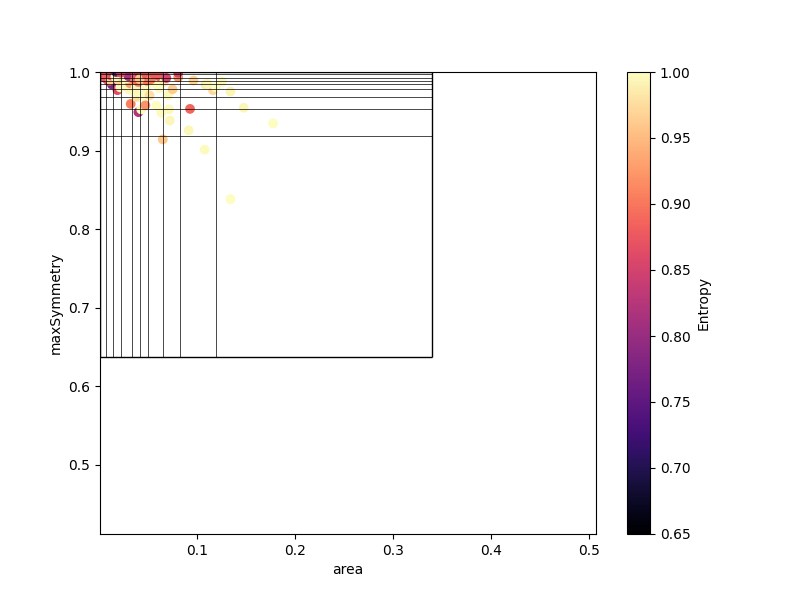}}

	\subfloat[Point-Line]{\includegraphics[width=0.48\textwidth]{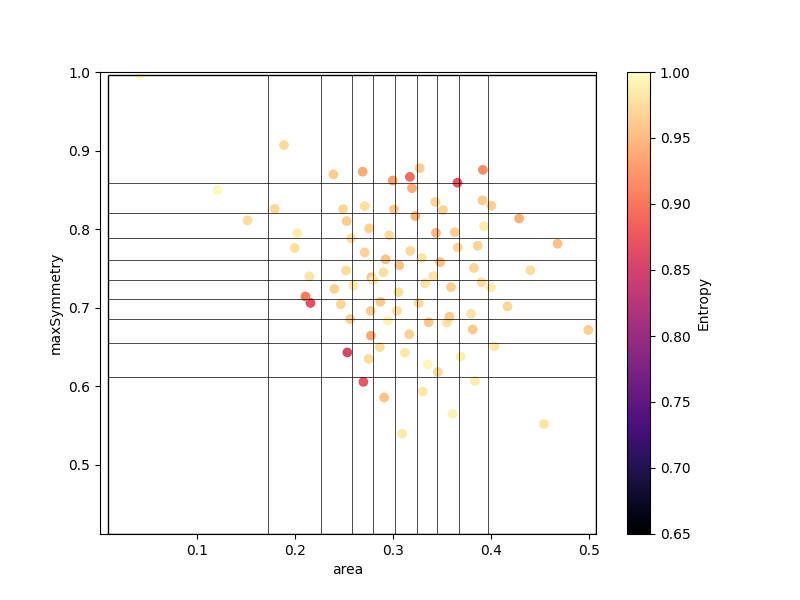}}
	\quad
	\subfloat[\SMT]{\includegraphics[width=0.48\textwidth]{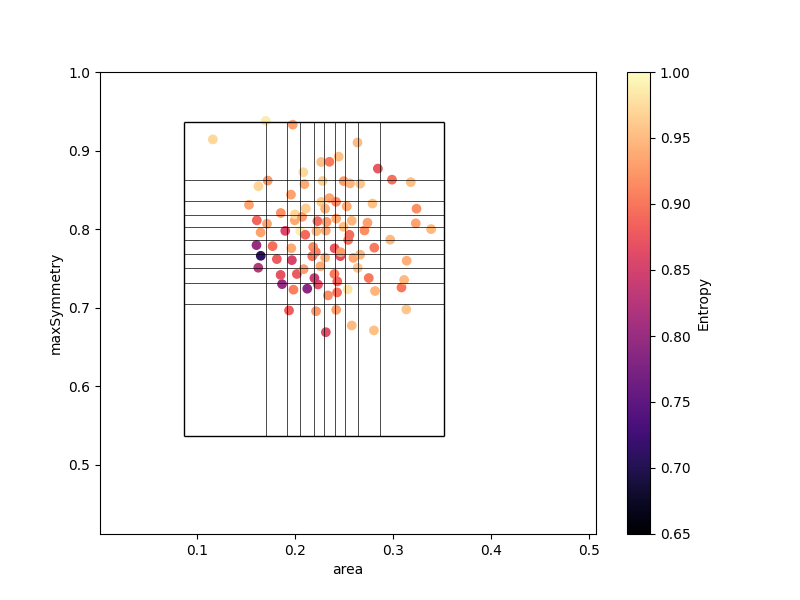}}
	
	\caption{Heatmap of the archives generated with the \textit{area} and \textit{maxSymmetry} features using the four genomes.}
	\label{fig:heatmap_exp1}
\end{figure}

	%
\begin{figure}[t]
	\subfloat[Max Entropy]{\includegraphics[width=0.48\textwidth]{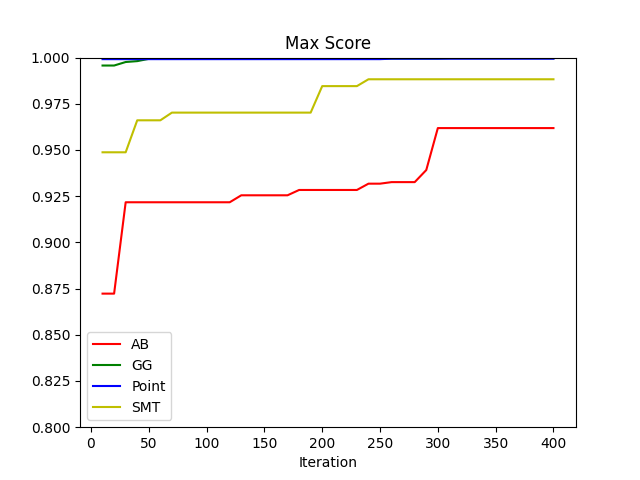}
	\label{fig:max_score_exp1}	
	}
	\quad
	\subfloat[CCDF]{\includegraphics[width=0.48\textwidth]{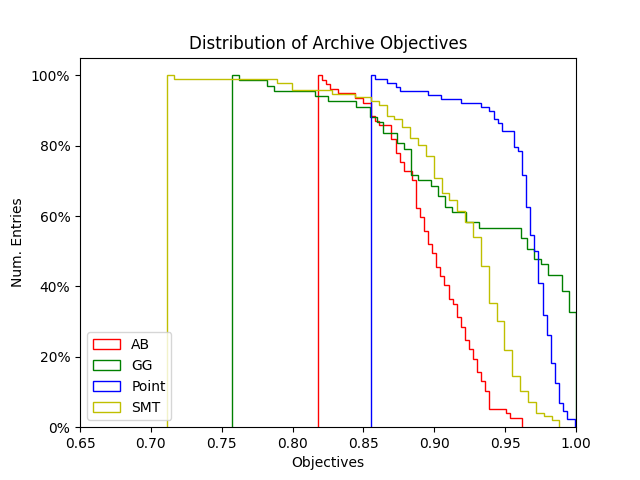}
		\label{fig:ccdf_exp1}	
	}

	%
	\subfloat[QD Score]{\includegraphics[width=0.48\textwidth]{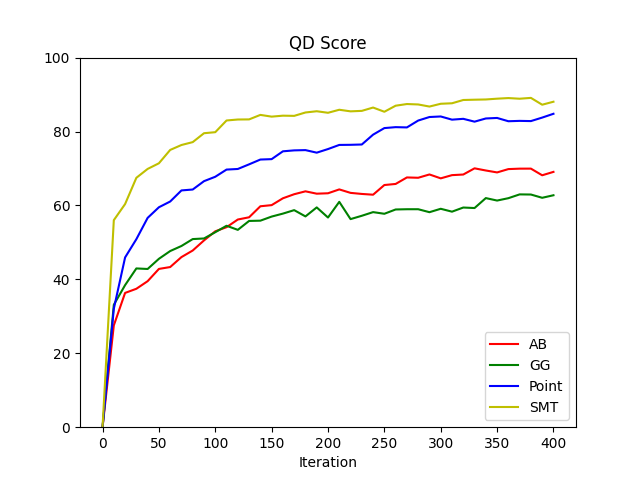}
		\label{fig:qd_score_exp1}	
	}
	\quad
	\subfloat[Size]{\includegraphics[width=0.48\textwidth]{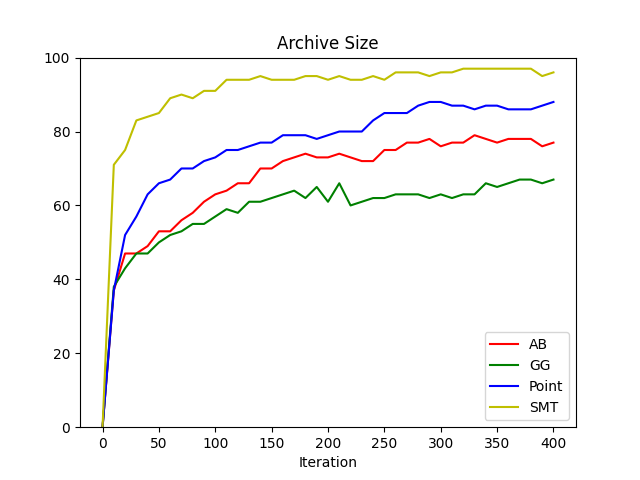}
	\label{fig:size_exp1}
}
	\caption{Experiments with \textit{area} and \textit{maxSymmetry} features: 
		(a) maximum entropy,
		(b) CCDF of the fitness of elites,
		(c) QD score of the archive,
		(d) size of the archive over time.		
	}
	\label{fig:area-maxsimmetry}
\end{figure}

%% file: sections/fig_visibility_exp1.tex
\begin{figure}[t]
	\centering
	
	\subfloat[]{
		\includegraphics[width=0.4\textwidth]{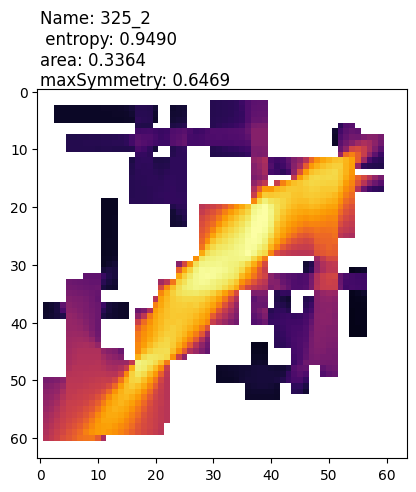}
		\label{fig:best_maps_exp1_ab1}		
	}
	\quad
	\subfloat[]{
		\includegraphics[width=0.4\textwidth]{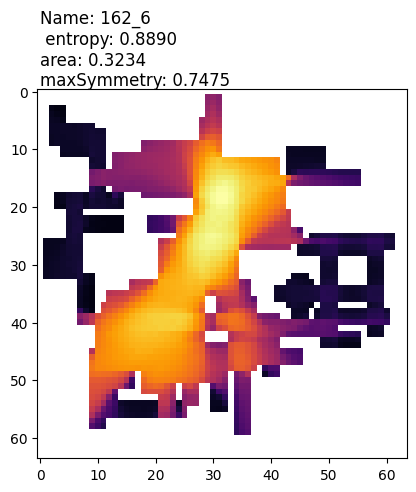}
		\label{fig:best_maps_exp1_ab2}
	}

	\subfloat[]{
		\includegraphics[width=0.4\textwidth]{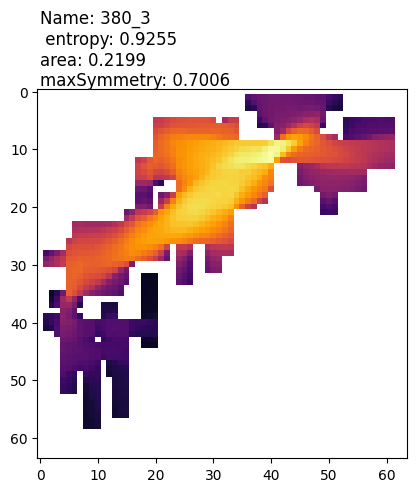}
		\label{fig:best_maps_exp1_ab3}		
	}
	\quad
	\subfloat[]{
		\includegraphics[width=0.4\textwidth]{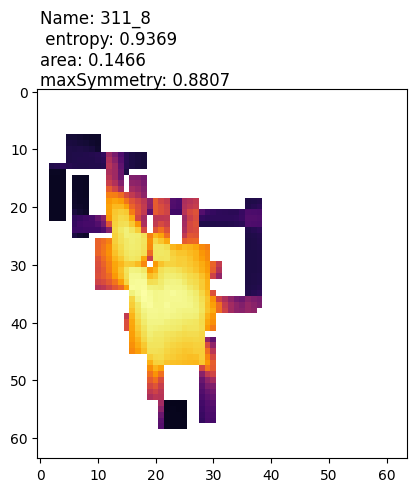}
		\label{fig:best_maps_exp1_ab4}		
	}
\caption{Best performing \textit{All-Black} maps in the archive obtained with the \textit{area} and \textit{maxSymmetry} features.}	
\label{fig:best_maps_ab_exp1}
\end{figure}

\begin{figure}[t]
	\centering	
	\subfloat[]{
	\includegraphics[width=0.4\textwidth]{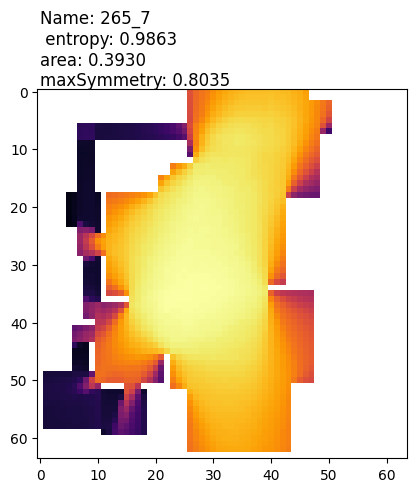}
				\label{fig:best_maps_exp1_pl1}
	}
	\subfloat[]{
		\includegraphics[width=0.4\textwidth]{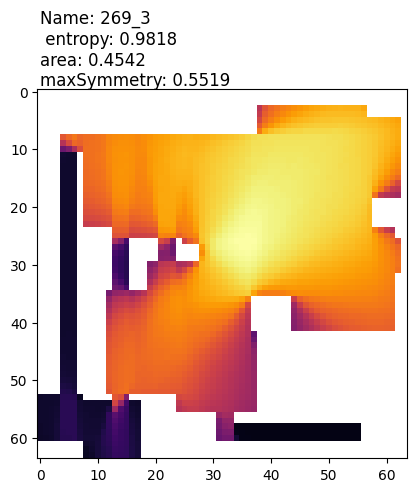}
					\label{fig:best_maps_exp1_pl2}
	}

	\subfloat[]{
		\includegraphics[width=0.4\textwidth]{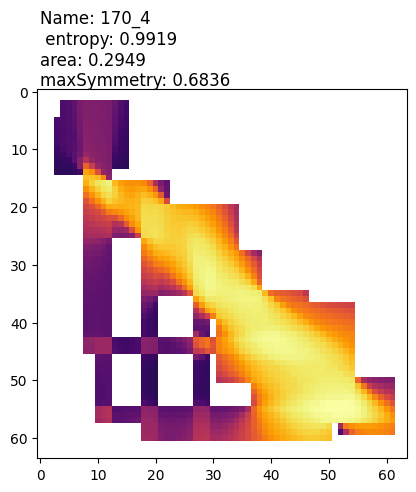}
		\label{fig:best_maps_exp1_pl3}
	}
	\subfloat[]{
		\includegraphics[width=0.4\textwidth]{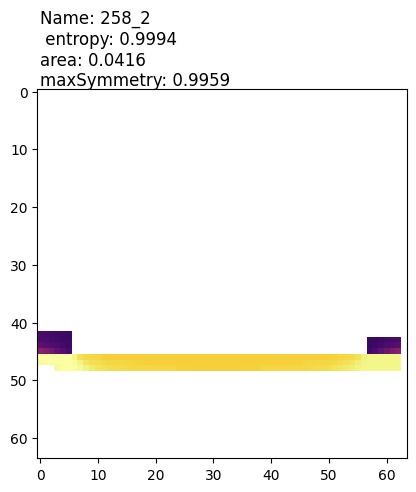}		
		\label{fig:best_maps_exp1_pl4}
	}
	
\caption{Best performing \textit{Point-Line} maps in the archive obtained with the \textit{area} and \textit{maxSymmetry} features.}
\label{fig:best_maps_point_exp1}
\end{figure}

\begin{figure}[t]
	\centering
	\subfloat[]{	
		\includegraphics[width=0.4\textwidth]{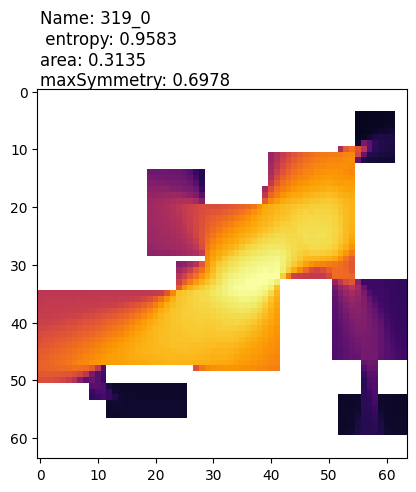}
		\label{fig:best_maps_exp1_smt1}
	}
	\quad
	\subfloat[]{
		\includegraphics[width=0.4\textwidth]{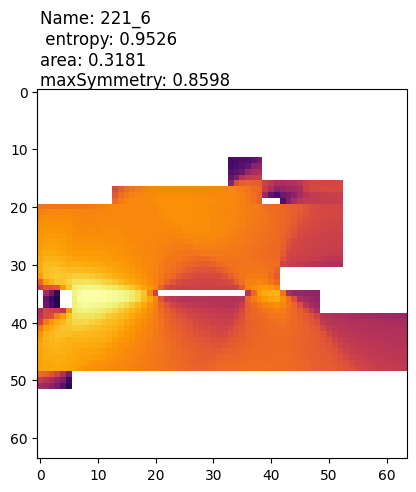}
		\label{fig:best_maps_exp1_smt2}
	}

	\subfloat[]{
		\includegraphics[width=0.4\textwidth]{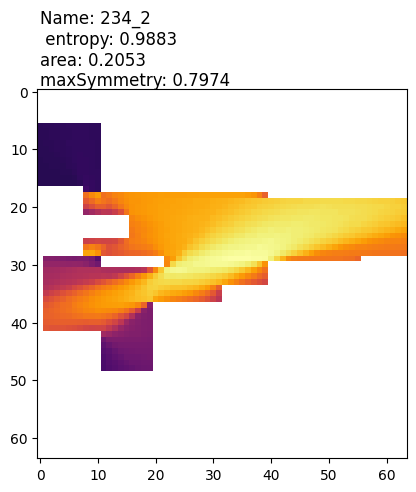}
		\label{fig:best_maps_exp1_smt3}
	}
	\quad
	\subfloat[]{
		\includegraphics[width=0.4\textwidth]{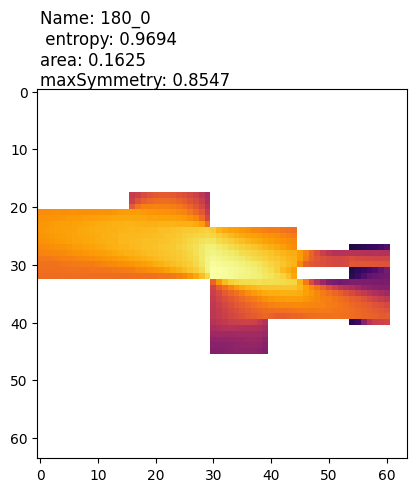}
		\label{fig:best_maps_exp1_smt4}
	}
	
	\caption{Best performing \SMT\ maps in the archive obtained with the \textit{area} and \textit{maxSymmetry} features.}
	\label{fig:best_maps_smt_exp1}
\end{figure}

%% file: sections/ssec_pace_average_eccentricity.tex
In the second set of experiments, we applied MESB illuminating the search space using \textit{pace} and \textit{averageEccentricity}. 
The heatmaps in Figure~\ref{fig:heatmap_exp2} show that \textit{Grid-Graph} generates fewer, sparser, bins, suggesting that the \textit{area}-\textit{maxSymmetry} space is more difficult to illuminate using this representation. \textit{Grid-Graph} maps also have lower \textit{averageEccentricity}, thus a more centralized layout, and higher \textit{pace}.
\SMT\ does not generate maps with sparser layouts as its \textit{averageEccentricity} values do not go beyond 120.

The results for the maximum fitness value (Figure~\ref{fig:max_score_exp2}) and the distribution of the fitness of elites (Figure~\ref{fig:ccdf_exp2})
	are similar to the previous experiments. \textit{Grid-Graph} and \textit{Point-Line} can achieve higher entropy values both on average and in the best maps, 
	while \textit{All-Black} and \SMT\ lag behind. 
The QD score (Figure~\ref{fig:qd_score_exp2}) shows that \textit{Grid-Graph} severely under-performs due to its under-filled archive, 
	while the other representations boast similar scores. 
%
%
Figure~\ref{fig:size_exp2} shows that \textit{Grid-Graph} has the fewest elites in the archive suggesting once again 
	that the representation is limited in the amount of variety it can generate. 
In contrast, 
	\textit{All-Black}, \textit{Point-Line} and \SMT\ cover the archive in similar ways, 
	although \SMT\ does it faster as already noted in the previous experiment (Figure~\ref{fig:size_exp2}).

\input{sections/fig_performance_exp2}
\input{sections/fig_visibility_exp2}

\medskip\noindent\textbf{Analysis of Maps in the Archive.} Figures \ref{fig:best_maps_exp2_ab}, \ref{fig:best_maps_exp2_pl}, and \ref{fig:best_maps_exp2_smt} shows examples of the best performing (high entropy, well-balanced) maps evolved; for each representation, the maps in the upper rows have a much higher pace than the maps in the bottom row; the maps on the left have a much lower \textit{averageEccentricity} than the ones on the right. 
As before, we did not report the maps evolved with the Grid-Graph representation since they are too simple and refer the interested reader to \cite{denodato:2024:thesis} for examples and more results.

The maps evolved using \textit{pace} and \textit{averageEccentricity} show a wider variety compared to the ones produced in the previous experiments and the features heavily influence the map topology. Higher pace and averageEccentricity values generate maps with more complex layouts; matches with lower pace values require alternate pathways with loops allowing bots to escape fighting (e.g., Figure~\ref{fig:best_maps_exp2_ab1}, \ref{fig:best_maps_exp2_pl1}, and \ref{fig:best_maps_exp2_smt1}). 
Increasing values of  \textit{averageEccentricity} lead to maps  with more centralized and thus more connected layouts, so that all rooms are easily reachable and closer to the map center (e.g., \ref{fig:best_maps_exp2_ab4}, \ref{fig:best_maps_exp2_pl4}, \ref{fig:best_maps_exp2_smt4}).

\textit{All-Black} arguably produces the most intricate maps 
with many short and long corridors, loops, covers and alternate pathways between rooms. 
However,  \textit{All-Black} layouts still show the well-known representation's issues \cite{olsted_interactive_2015} involving noisy layouts, many dead ends, and useless features, which are hardly utilized during gameplay. This suggests that simply using the \textit{entropy} as fitness will often lead to maps that are balanced \textit{in principle} but that, in reality, do not apply basic design principles that all good FPS maps should have. 
%
%
In contrast, quality diversity, coupled with our representations (Point-Line and \SMT), 
	generate the best trade-off between balance and design, as opposed to previous optimization algorithms (e.g., \cite{cardamone_evolving_2011,lanzi_evolving_2014,loiacono_fight_2017}), which would only find the best maps in the feature space.
Our representations generate cleaner maps that also enable several gameplay tactics either with long loops and long corridors to achieve low \textit{pace} values, or very centralized maps with a big central room to achieve high pace gameplay.

%% file: sections/fig_performance_exp2.tex
\begin{figure*}[t]
	\centering
	\subfloat[All-Black]{\includegraphics[width=0.48\textwidth]{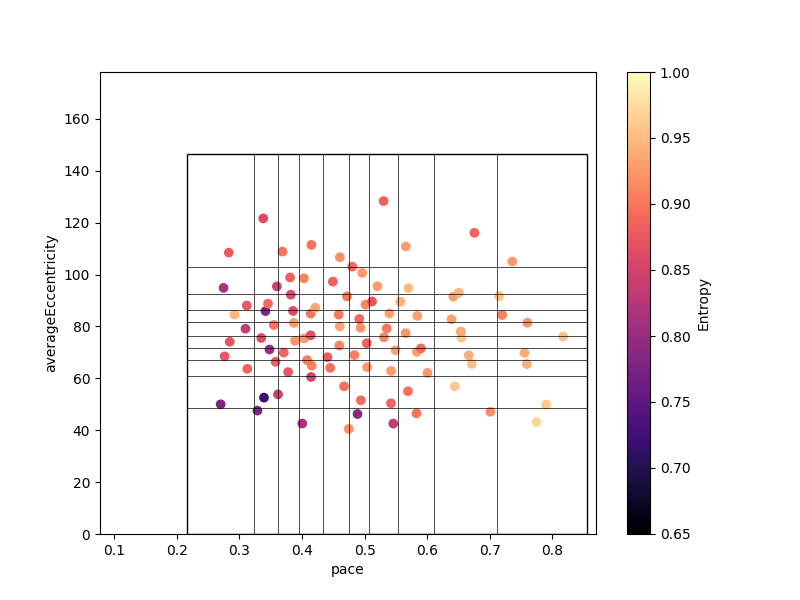}}
	\quad
	\subfloat[Grid-Graph]{\includegraphics[width=0.48\textwidth]{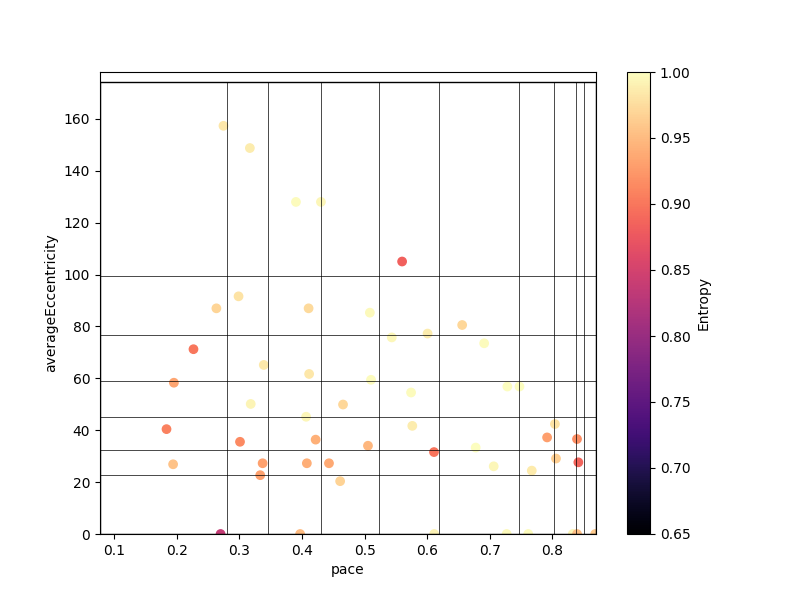}}

	\subfloat[Point-Line]{\includegraphics[width=0.48\textwidth]{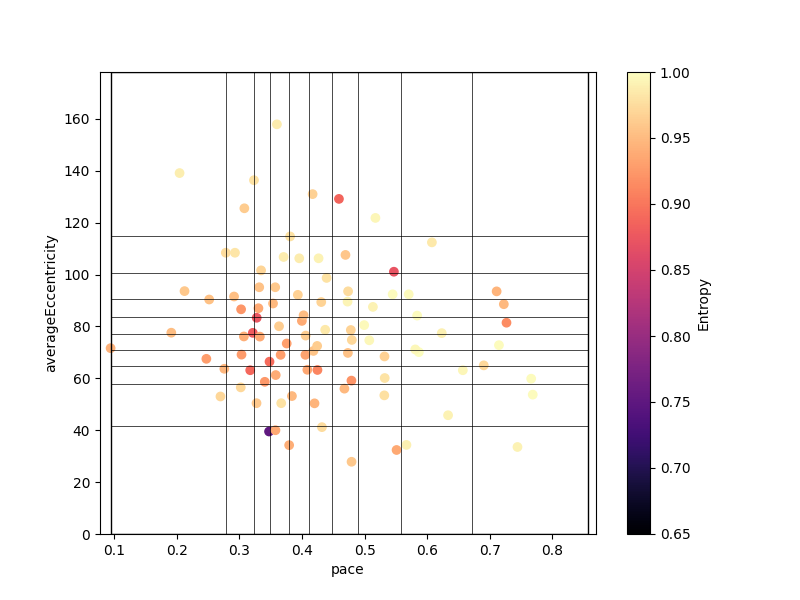}}
	\quad
	\subfloat[Spatial Layout]{\includegraphics[width=0.48\textwidth]{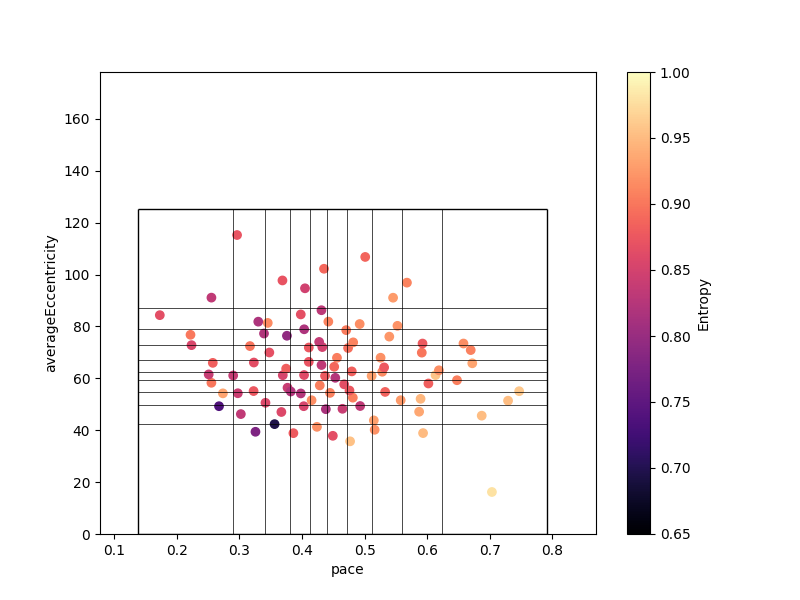}}
	\caption{Heatmap of the archives generated with the \textit{pace} and \textit{averageEccentricity} features using the four genomes.}	
	\label{fig:heatmap_exp2}
\end{figure*}

\begin{figure*}	
	\centering
	\subfloat[Max Entropy]{\includegraphics[width=0.48\textwidth]{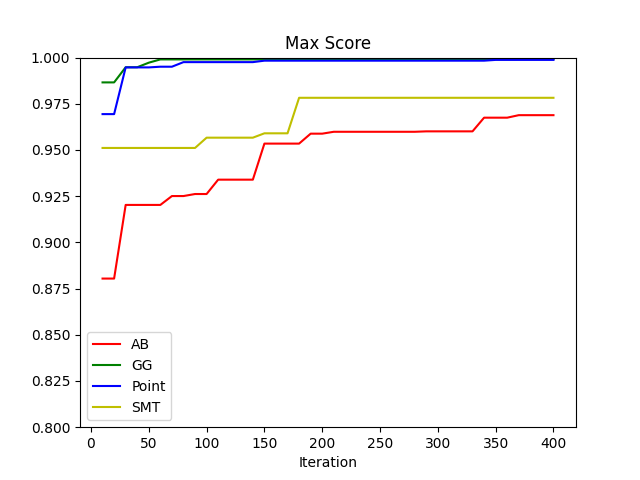}
		\label{fig:max_score_exp2}	
	}
	\subfloat[CCDF]{\includegraphics[width=0.48\textwidth]{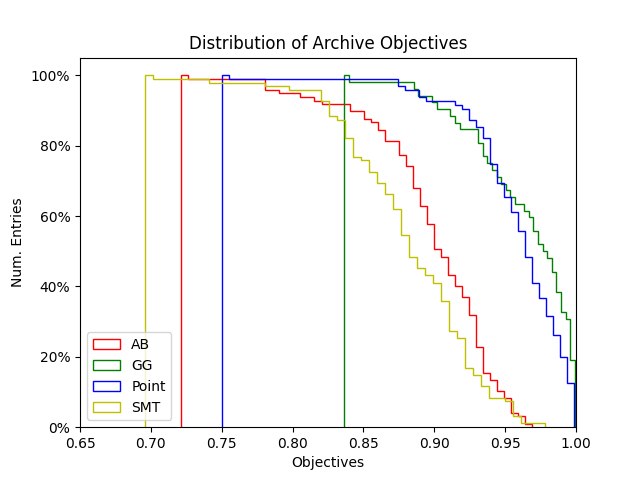}
		\label{fig:ccdf_exp2}	
	}
	
	\subfloat[QD Score]{\includegraphics[width=0.48\textwidth]{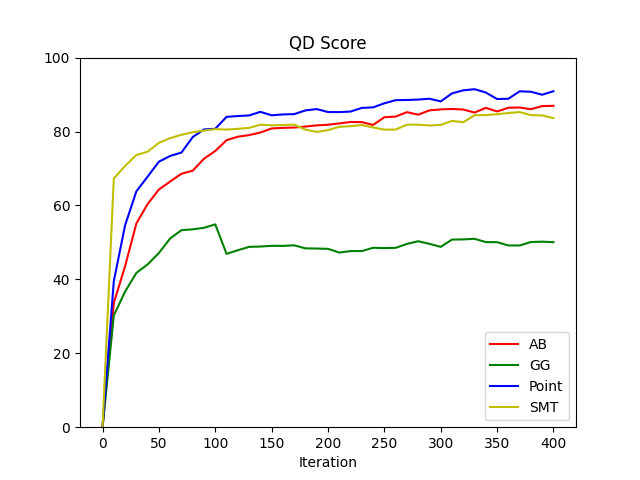}
		\label{fig:qd_score_exp2}	
	}
	\subfloat[Size]{\includegraphics[width=0.48\textwidth]{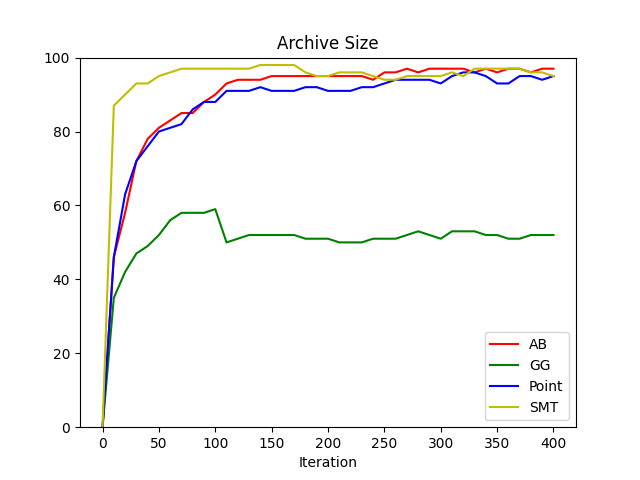}
		\label{fig:size_exp2}
	}
	\caption{Experiments with \textit{pace} and \textit{averageEccentricity} features: 
		(a) maximum entropy,
		(b) CCDF of the fitness of elites,
		(c) QD score of the archive,
		(d) size of the archive over time.
	}
	\label{fig:pace-average-eccentricity}
\end{figure*}

%

%% file: sections/fig_visibility_exp2.tex
\begin{figure*}
			\centering
			\subfloat[]{
				\includegraphics[width=0.48\textwidth]{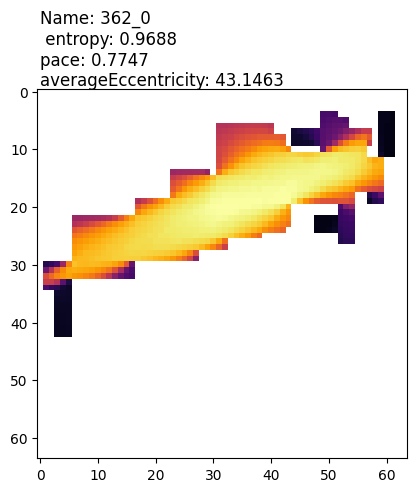}
				\label{fig:best_maps_exp2_ab3}
			}
			\subfloat[]{
				\includegraphics[width=0.48\textwidth]{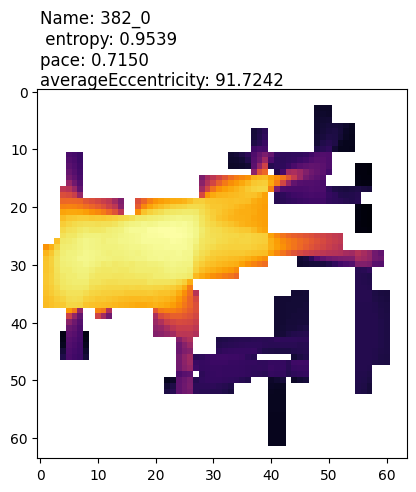}
				\label{fig:best_maps_exp2_ab4}
			}
			
			\qquad
			\subfloat[]{
				\includegraphics[width=0.48\textwidth]{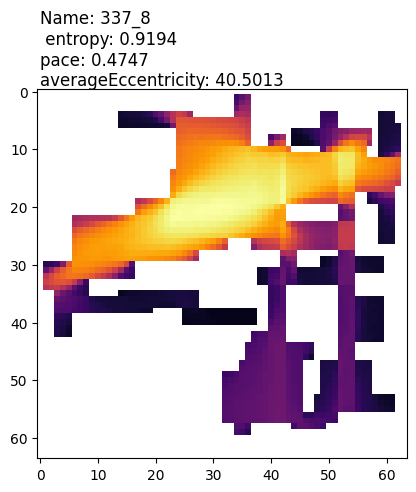}
				\label{fig:best_maps_exp2_ab1}
			}
			\subfloat[]{
				\includegraphics[width=0.48\textwidth]{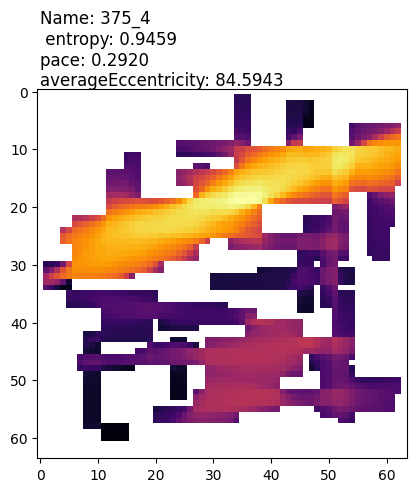}
				\label{fig:best_maps_exp2_ab2}
			}
	\caption{Best performing maps in the archive evolved with the \textit{All-Black} representation using \textit{pace} and \textit{averageEccentricity}. 
		Maps are shown using visibility matrices in which lighter colors identify more visible areas.}	
	\label{fig:best_maps_exp2_ab}			
\end{figure*}

\begin{figure*}
	\centering
			\subfloat[]{
				\includegraphics[width=0.48\textwidth]{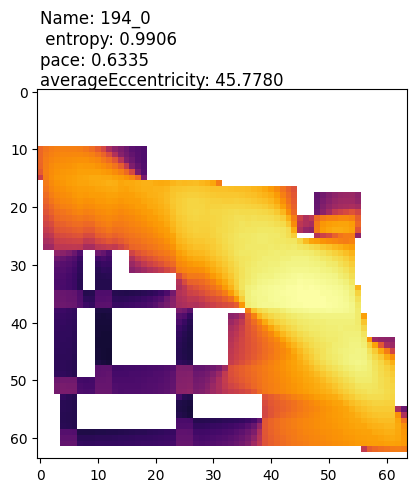}
				\label{fig:best_maps_exp2_pl3}
			}
			\subfloat[]{
				\includegraphics[width=0.48\textwidth]{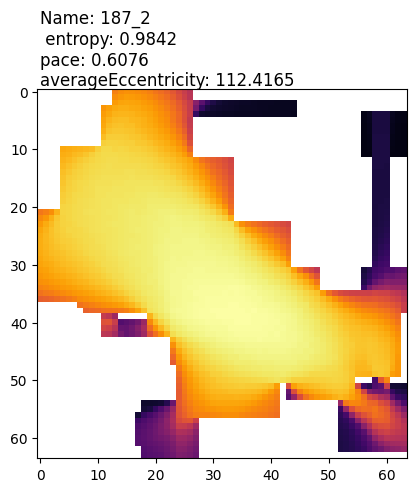}		
				\label{fig:best_maps_exp2_pl4}
			}
			
			\qquad
			\subfloat[]{
				\includegraphics[width=0.48\textwidth]{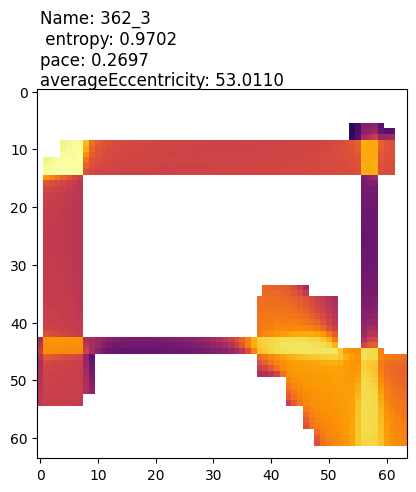}
				\label{fig:best_maps_exp2_pl1}
			}
			\subfloat[]{
				\includegraphics[width=0.48\textwidth]{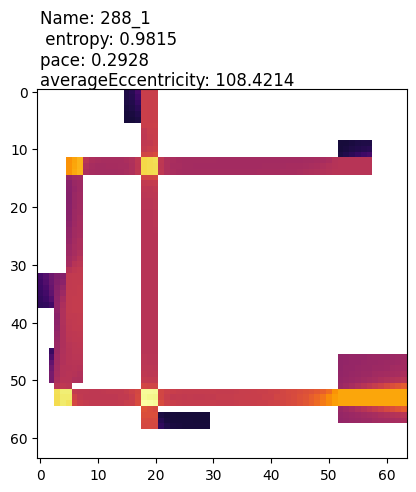}
				\label{fig:best_maps_exp2_pl2}
			}		
			
	\caption{Best performing maps in the archive evolved with the \textit{Point-Line} representation using \textit{pace} and \textit{averageEccentricity}. 
	Maps are shown using visibility matrices in which lighter colors identify more visible areas.}	
\label{fig:best_maps_exp2_pl}			
\end{figure*}

\begin{figure*}		
	\centering
	\subfloat[]{
		\includegraphics[width=0.48\textwidth]{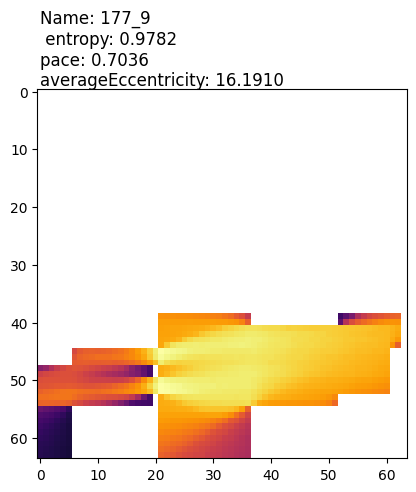}
		\label{fig:best_maps_exp2_smt3}
	}
	\subfloat[]{
		\includegraphics[width=0.48\textwidth]{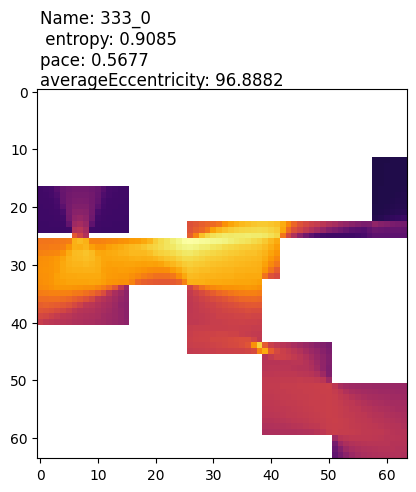}
		\label{fig:best_maps_exp2_smt4}
	}
	
	\qquad
	\subfloat[]{	
		\includegraphics[width=0.48\textwidth]{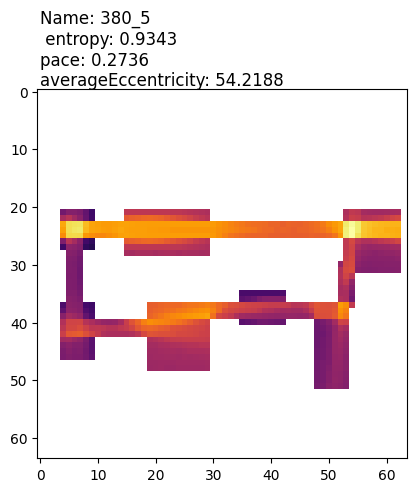}
		\label{fig:best_maps_exp2_smt1}
	}
	\subfloat[]{
		\includegraphics[width=0.48\textwidth]{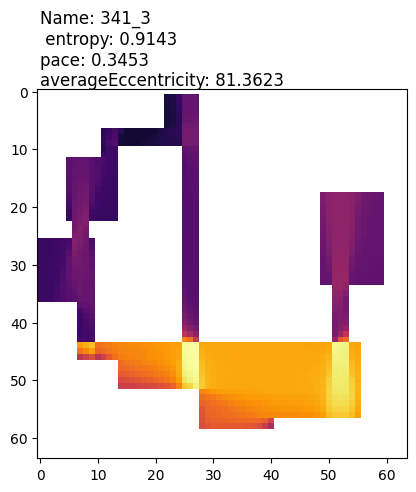}
		\label{fig:best_maps_exp2_smt2}
	}
			
	\caption{Best performing maps in the archive evolved with the \textit{\SMT}\ representation using \textit{pace} and \textit{averageEccentricity}. 
	Maps are shown using visibility matrices in which lighter colors identify more visible areas.}	
	\label{fig:best_maps_exp2_smt}			
\end{figure*}

%

%% file: sections/sec_conclusions.tex
We investigated the application of MAP-Elites with Sliding Boundaries (MESB) to the evolution of maps for First-Person Shooter (FPS) games. We introduced two novel representations (Point-Line and \SMT) we designed to tackle the limitations of existing well-known representations of FPS maps \cite{cardamone_evolving_2011,lanzi_evolving_2014,olsted_interactive_2015}. 
We defined several metrics to describe the maps’ topological and emergent properties. We selected the features that, according to the literature, are considered the most relevant for FPS map design \cite{DBLP:conf/fdg/HullettW10,cardamone_evolving_2011,DBLP:conf/cig/BallabioL19}. We combined the selected features into two pairs (\textit{area-maxSymmetry} and \textit{pace-averageEccentricity}) that we used to illuminate MAP-Elites.
We presented the results of two sets of experiments aimed at evolving balanced FPS maps using different pairs of features to illuminate the search space (\textit{area-maxSymmetry} and \textit{pace-averageEccentricity}). The results show that the combination of quality diversity and our representations (Point-Line and \SMT) generate the best trade-off between balance and design, unlike previous optimization algorithms and representations \cite{cardamone_evolving_2011,lanzi_evolving_2014,loiacono_fight_2017}. Our representations generate cleaner maps that enable several gameplay tactics, either with long loops and long corridors to achieve low-paced gameplay or very centralized maps with a large central room to achieve high-paced gameplay.
Our study is limited in that we used bots using only two weapons enabling long-distance and close combat fighting tactics, which, however, implement the most typical behaviors of human players \cite{hullett_design_2010,cardamone_evolving_2011}. 
Traditional search methods tend to rapidly focus on interesting (high-performing) maps that users can enjoy playing \cite{cardamone_evolving_2011,lanzi_evolving_2014, olsted_interactive_2015}. 
In contrast, for its nature, quality diversity methods explore a much wider range of maps, including the ones that, although enabling balanced gameplay, would be dull or uninteresting to play for human subjects. Accordingly, we believe that simulations using bots to implement stereotypical human strategies are the best option for our scenario.  